\documentclass[reprint, amsmath,amssymb,aps,showkeys]{revtex4-1}
\usepackage{subfigure}
\usepackage{graphicx}
\usepackage{dcolumn}
\usepackage{bm}
\usepackage{lineno}

\begin{document}
\preprint{APS/123-QED}

\title{Linking Gaussian Process regression with data-driven manifold embeddings\\ for nonlinear data fusion}

\author{Seungjoon Lee}
\affiliation{Department of Chemical and Biomolecular Engineering, Johns Hopkins University}

\author{Felix Dietrich}
\affiliation{Department of Chemical and Biomolecular Engineering, Johns Hopkins University}

\author{George E. Karniadakis}
\affiliation{Division of Applied Mathematics, Brown University}

\author{Ioannis G. Kevrekidis}
\affiliation{Department of Chemical and Biomolecular Engineering, Johns Hopkins University
}%
\email{yannisk@jhu.edu}
\altaffiliation[Also at ]{Department of Applied Mathematics and statistics and Department of Medicine, Johns Hopkins University.}


\begin{abstract}
In statistical modeling with Gaussian Process regression, it has been shown that combining (few) high-fidelity data with (many) low-fidelity data can enhance prediction accuracy, compared to prediction based on the few high-fidelity data only.
Such information fusion techniques for multifidelity data commonly approach the high-fidelity model $f_h(t)$ as a function of {\emph two} variables $(t,y)$, and then using $f_l(t)$ as the $y$ data.
More generally, the high-fidelity model can be written as a function of several
variables $(t,y_1,y_2....)$; the low-fidelity model $f_l$ and, say, some of its derivatives, can then be substituted for these variables. 
In this paper, we will explore mathematical algorithms for multifidelity information fusion that use such an approach towards improving the representation of the high-fidelity function with only a few training data points.
Given that $f_h$ may not be a simple function -and sometimes not even a function- of $f_l$, we demonstrate that using additional functions of $t$, such as derivatives or shifts of $f_l$, can drastically improve the approximation of $f_h$ through Gaussian Processes.
We also point out a connection with ``embedology" techniques from topology and dynamical systems.
%
%
Our illustrative examples range from instructive caricatures to computational biology models, such as Hodgkin-Huxley neural oscillations.


\end{abstract}

\pacs{Valid PACS appear here}
\maketitle

\section{Introduction}

\begin{figure*}
 \centering
  \subfigure[A correlation between $t$, $f_l(t)$, and $f_h(t)$]{
  \includegraphics[scale=0.3]{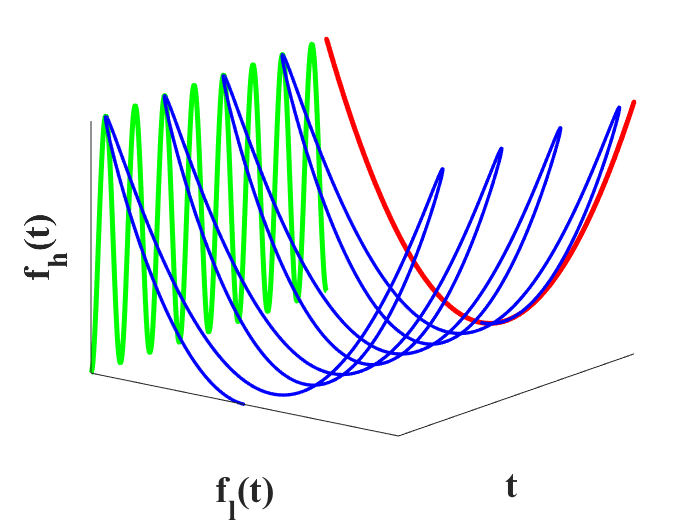}
   }
 \subfigure[GP with \{t\}]{
  \includegraphics[scale=0.35]{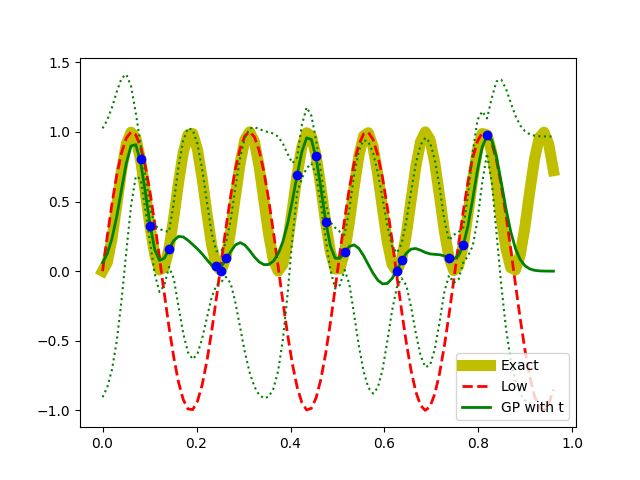}
   }
 \subfigure[GP with \{$f_l(t)$\}]{
  \includegraphics[scale=0.35]{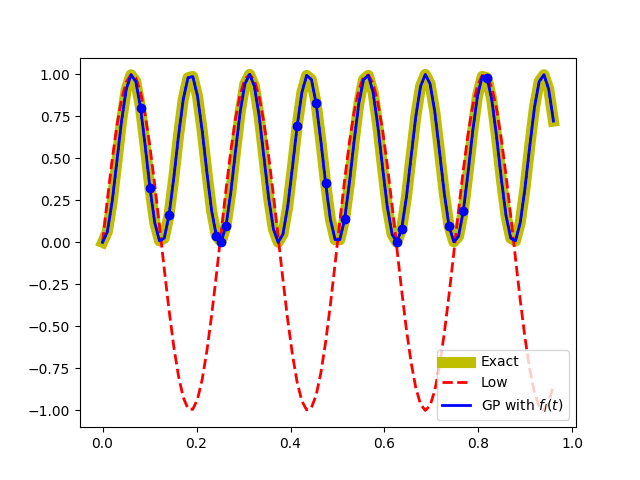}
   }

  \caption{Two alternative GP regressions for the high-fidelity model $f_h(t)=\sin^2({8\pi t})$ (the highly oscillatory green curve in (a)).
  The blue curve visually suggests a smooth two-dimensional manifold $g(t,f_l)$.   
  If, on the other hand, the high-fidelity function is projected onto $f_l(t)$, we can see it is a simple quadratic. 
  (b) and (c): the prediction results with 2 standard deviations (dashed line) for the high-fidelity function by GP regression with only $t$ and with only $f_l(t)$, respectively. We employ 15 high-fidelity data points and 100 uniformly distributed low-fidelity data points for the regression.
  } 
\label{fig:simple}
\end{figure*}

Recent advances in both algorithms and hardware are increasingly making machine learning an important component of mathematical modeling for physicochemical, engineering, as well as biological systems (e.g. ~\cite{Tarca07,Holzinger14,Libbrecht15,Villoutreix17}).
Part of these developments focus on multiresolution and multifidelity data fusion~\cite{Lanckriet04,Willett13}.
Fusing information from models constructed at different levels of resolution/fidelity has been shown to enhance prediction accuracy in data-driven scientific computing (e.g.~\cite{lee17R,lee17D}).

In addition, if high-fidelity data are costly to obtain (experimentally or computationally) while low-fidelity data are relatively cheap, a combination of a few high-fidelity data and many low-fidelity data can also lead to overall computational efficiency.
Recently, Gaussian Process (GP) regression has been widely used to effectively combine multiple fidelity data~\cite{le14,perdikaris15,perd16,parussini17,perdikaris17}.
``Classical" information fusion by GP had focused only on  linear correlations between high- and low-fidelity data via auto-regressive schemes such as the coKriging approach of Kennedy and O'Hagan~\cite{kennedy00}.
If two (or more) data sets have {\em nonlinear} correlations, such linear-type approaches will lose their effectiveness.  
   
When a high-fidelity model $f_h(t)$ is nonlinearly correlated with a low-fidelity model $f_l(t)$, Nonlinear Auto-Regressive Gaussian Process (NARGP)~\cite{perdikaris17} achieves highly accurate prediction results by introducing an additional dimension: $f_h(t)$ is approximated as a curve on a two-dimensional manifold parametrized by $t$ and a second variable, $y$. The data points on this manifold are of the form $f_h(t), t, f_l(t)$.
More generally, NARGP finds a strong correlation between $d$-dimensional high- and low-fidelity models in a $(d+1)$-dimensional space.
In this framework, the low-fidelity model provides the additional ``latent" variable of the high-fidelity model.

Deep multifidelity GPs were introduced~\cite{raissi16} as an improvement, especially in the context of discontinuities in $f_h$ with respect to $t$ (and $f_l$).
This approach focuses on constructing a useful transformation $T(t)$ of the input variables $t$ through a (deep) neural network.
Then, the high-fidelity function $f_h(t)$ is approximated as a GP of (just) $f_l(T(t))$.
One must now, of course, perform optimization for the additional network hyperparameters.
    
In this paper we discuss a connection of NARGP with data-driven embeddings in topology/dynamical systems, and extend it (in the spirit of such data driven embeddings) in an attempt to improve the numerical approximation of $f_h$ in the ``sparse $f_h$, rich $f_l$" data setting.
In what follows we will (rather arbitrarily) ascribe the characterization ``high fidelity" or ``low fidelity" to different functions used in our illustrations; this characterization is solely based on the number of available data points for each.

For our first example, the ``high-fidelity" function $f_h$, for which we only have a few data points, is a function of $t$; but it actually also happens 
that we can describe it as a function of $f_l$, 
\begin{equation}
f_h(t) = \sin^2 (8\pi t), \;\;\;\; f_l(t) = \sin (8\pi t),
\end{equation}
see FIG.\ref{fig:simple}.

When we choose $t$ as the coordinate parametrizing $f_h(t)$ (the green curve in FIG.\ref{fig:simple}(a)), the GP regression fails to represent the high-frequency sine function with just a few training data points as shown in FIG.\ref{fig:simple}(b). 
However, as FIG.\ref{fig:simple}(c) shows, if we choose $f_l$ as the coordinate of $f_h(f_l)$ (colored by red in FIG.\ref{fig:simple}(a)), the GP regression can represent the simple quadratic function quite effectively.
If we still need to know the parametrization of $f_h$ by $t$,
we can obtain it through the ``data rich" $f_l$:  $f_h(t) \equiv f_h(f_l(t))$.

If $f_h$  {\emph is not} a function of $f_l$, however (as has been observed in the literature~\cite{perdikaris17} and as can be seen in Figure \ref{fig:embed}(a)) more variables are necessary to create a useful parametrization of $f_h$ - more precisely, a domain over which $f_h$ is a function. 

In the NARGP framework, the variable used in addition to $f_l$ is $t$ itself.  In this paper, we will also advocate the use of delays or derivatives of $f_l$ as additional variables.
This approach can also help remove the explicit dependence of $f_h$ on $t$, since embedding theories in dynamical systems~\cite{Whitney36, Nash66, Takens81, kostelich90, sauer91,kennel92} guarantee that we can choose any generic observation of $t$, or derivatives and/or delays of this observation,  as a replacement for $t$, see section \ref{sec:embed} for more details.

In section \ref{sec:HH}, we apply the proposed framework to the Hodgkin-Huxley model, describing the behavior of action potentials in a neuron.
Here, the high- and the low-fidelity functions are action potentials at two different values of the external current.
This is a case where $f_h$ is a complicated function of $t$
and does not only depend on $f_l$; yet as we will see, delays of $f_l$ will help us construct an accurate approximation of $f_h$.

The paper is organized as follows: In section \ref{sec:method}, we review the NARGP framework and concepts of ``embedology".
Also, we illustrate why and when this framework is successful.
In section \ref{sec:result}, we demonstrate the effectiveness of our proposed framework via some pedagogical examples and the biologically motivated application to the Hodgkin-Huxley model.
In section \ref{sec:conclusion}, we summarize our results and discuss open issues for further development of general multifidelity information fusion in modeling and simulation practice across scientific domains.

\section{Methods}
\label{sec:method}

\subsection{Nonlinear information fusion algorithms}

\begin{figure*}
 \centering
  \subfigure[A correlation between $t$, $f_l(t)$, and $f_h(t)$]{
  \includegraphics[scale=0.33]{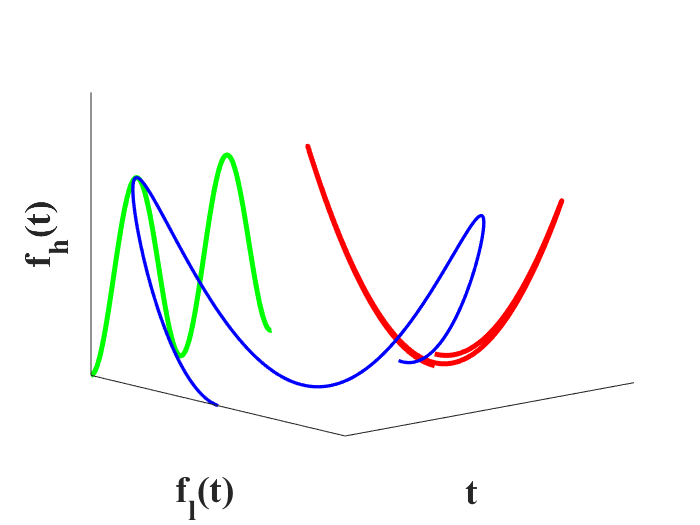}
   }
 \subfigure[NARGP: GP with \{$t$, $f_l(t)$\}]{
  \includegraphics[scale=0.33]{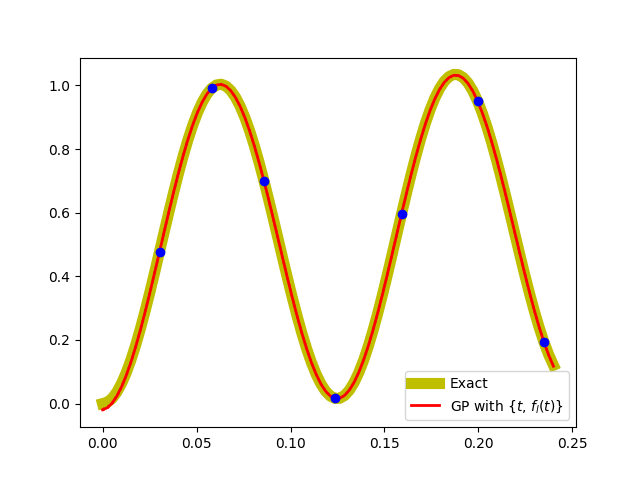}
   }
 \subfigure[GP with \{$f_l(t)$, $f_l(t-\tau)$\}]{
  \includegraphics[scale=0.33]{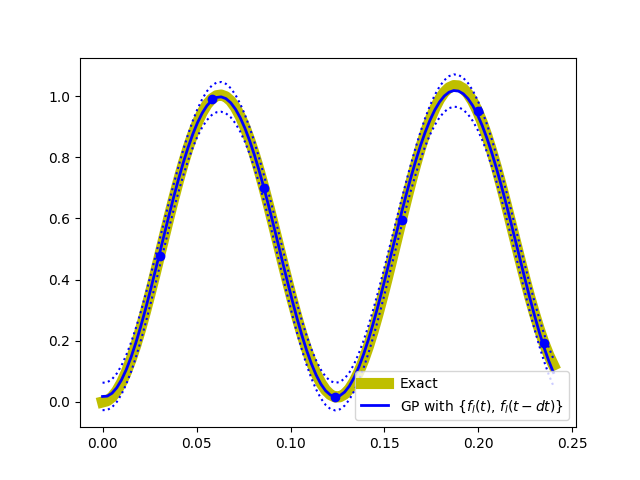}
   }
  \caption{Two alternative GP regressions for the high-fidelity model $f_h$ (see Equation (\ref{eqn:embed})). (a): the multivaluedness of $f_h$ with respect to $f_l$ is clearly visible when the high-fidelity data (the blue curve) is projected onto $f_l(t)$ (the red curve). (b) and (c): The high-fidelity function (the yellow curve) versus posterior means with two standard deviations (dashed lines) of two alternative GP regressions. We use 7 high-fidelity data points and 100 uniformly distributed low-fidelity data points for training GP regression models. (b) GP regression with \{$t$, $f_l(t)$\}. (c) GP regression with \{$f_l(t)$, $f_l(t-\tau)$\}, $\tau=1/400$.} 
\label{fig:embed}
\end{figure*}

``Classical" multifidelity data fusion algorithms require a linear (or almost linear) correlation between different fidelity models.
Under this constraint, we can merge two or more data sets by using scaling and shifting parameters such as the Kennedy and O'Hagan coKriging approach~\cite{kennedy00}.
However, more generally, the data sets are nonlinearly correlated, which typically degrades the quality of results of linear information fusion algorithms.
In order to resolve correlation nonlinearity between data sets,  the use of a space-dependent scaling factor $\rho(x)$~\cite{le13} or, alternatively, deep multifidelity Gaussian Processes~\cite{raissi16} have been introduced; clearly, the improvement they bring requires additional hyperparameter optimization.

When the high-fidelity model $f_h$ is nonlinearly correlated with the low-fidelity model $f_l$, but can be written as a simple function of $x$ and $f_l$, the Nonlinear Auto-Regressive GP (NARGP)~\cite{perdikaris17} is an appropriate choice.
In this framework, a one dimensional high-fidelity function $f_h$ is assumed to be a ``simple" function $g$ of two variables $(x, y)$, i.e. it is a curve that lies in the two-dimensional manifold described by $g$.
Then, GP regression in the two dimensional space is performed, where the data for $y$ are the $f_l(x)$ data,
\begin{equation}
g(x,y) \sim GP \left(0, k((x,y), (x',y'))\right), \; f_h(x) = g(x,f_l(x)).
\end{equation}
In ~\cite{perdikaris17} the gain in accuracy of NARGP, compared to an auto-regressive scheme with a constant scaling factor, as well as a scaling factor that was modeled as a (space-dependent) Gaussian process, was documented.

Algorithmically, classical autoregressive GPs employ an {\em explicit} method such as a scaling constant ($\rho$) between two covariance kernels, $k_1$ and $k_2$ as
\begin{equation}
\begin{bmatrix}
f_l(x) \\ f_h(x)
\end{bmatrix} \sim GP
\left(
\begin{bmatrix}
0 \\ 0
\end{bmatrix},
\begin{bmatrix}
k_1(x,x') & \rho k_1(x,x') \\ \rho k_1(x,x') & \rho^2k_1(x,x') + k_2(x,x')
\end{bmatrix}
\right).
\end{equation}

The NARGP framework, on the other hand, employs an {\em implicit} approach by the automatic relevance determination (ARD) weight~\cite{rasmussen06} in the extended space parametrized by $x$ and $y$: a different scaling hyperparameter for each of the two dimensions in the kernel.
In many applications, the radial basis function (see the equation (\ref{eqn:kernel})) has been used as the covariance kernel (where ARD implies a different scaling hyperparameter $\theta_i$ for each dimension):
\begin{equation}
k(x,x';\mathbf{\theta}) = \exp \left( -\frac{1}{2} \sum^d \theta_i (x_i - x'_i)^2 \right).
\label{eqn:kernel}
\end{equation}  

FIG.\ref{fig:embed} showcases an example where $f_h$ cannot be written as a function of $f_l$:
\begin{equation}
f_h(t) = t^2 + \sin^2 (8\pi t), \;\;\; f_l(t) = \sin (8\pi t) \;\;\; t\in[0, 0.25].
\label{eqn:embed}
\end{equation}
Following the NARGP framework, we choose the low-fidelity data as an additional variable $y=f_l(t)$; we then approximate the two-dimensional function
\begin{equation}
g(t,y) = t^2 + y^2.
\end{equation}
Approximating $g$ only requires a few training data point pairs for the GP regression. 
Then, $f_h$ can be written as $f_h(t) = g(t, f_l(t))$, see FIG.\ref{fig:embed}(b).
FIG.\ref{fig:embed}(c) demonstrates that we can, alternatively, use delays of $f_l$ instead of $t$ as an additional variable.
A rationalization of this follows in the next section.

\subsection{Data-driven Higher-dimensional Embeddings}
\label{sec:embed}

The theorem of Whitney~\cite{Whitney36} states that any sufficiently smooth manifold of dimension $d\in\mathbb{N}$ can be embedded in Euclidean space $\mathbb{R}^n$, with the tight bound $n\geq 2d+1$. Nash~\cite{Nash66} showed that this embedding can even be isometric if the manifold is compact and Riemannian, even though the bound on $n$ is higher. 
Many results on the reconstruction of invariant sets in the state spaces of dynamical systems are based on these two theorems~\cite{Takens81, kostelich90, sauer91,kennel92}. Here, the $n$ embedding dimensions are usually formed by $n$ scalar observations of the system state variables.
Instead of $n$ different observations, recording $n$ time delays of a single scalar observable is also possible, as originally formulated by Takens~\cite{Takens81} (see also~\cite{Packard80}.)

Given a smooth, $d$-dimensional manifold $M$, a vector field $X$ on $M$ with associated flow map $f:M\times \mathbb{R}\to M$, as well as an observable $h:M\to\mathbb{R}$, it is possible to construct an embedding $\phi:M\to\mathbb{R}^n$ through
\begin{equation}
\phi(p)=\left[h(p), h(f(p,-\Delta t)), \dots, h(f(p,-(n-1)\Delta t))\right]^T.
\end{equation}
Sauer et al.~\cite{sauer91} further specified the conditions on the observable, with respect to the underlying vector field of the dynamical system, that have to be satisfied such that the state space can be embedded successfully. They also extended the results on embeddings of invariant sets with fractal dimension. 
The observable and the vector field together build a tuple that has to be generic (Takens~\cite{Takens81}) or prevalent (Sauer et al.~\cite{sauer91}). The two papers show that ``almost all'' tuples are admissible, where the notions of genericity and prevalence are defining that probabilistic statement in the respective function spaces. These results are crucial for the numerical exploitation of the embedding theorems, since they show that ``almost all'' observables of a dynamical system can be chosen for a delay embedding. 

In many applications, the intrinsic dimension $d$ of $M$ is unknown, and $n$ is preemptively chosen large (larger than necessary). Manifold learning techniques are then often capable of reducing the embedding dimension, bringing it closer to the minimal embedding dimension necessary for the given manifold.

We also note that in our framework, if we can obtain low-fidelity data from a (dynamic) process instead of just single measurements, the same embedding and manifold learning techniques can be used even if the ``independent variable" $t$ is not known~\cite{Dietrich18}.



Let us consider the equation (\ref{eqn:embed}) again.
NARGP performs GP regression with two observations $\{t,f_l(t)\}$.
Using embedding theory, we can rationalize (a) why the two observations were necessary; and (b) why 
performing GP regression with an additional delay of the scalar observable, $\{f_l(t), f_l(t-\tau)\}$, is equally appropriate for a relatively small time horizon, see FIG.\ref{fig:embed}(c).
Notice that delay coordinates $f_l(t)$ and $f_l(t-\tau)$ lie on an ellipse with period of 0.25.
Hence, if the data are collected over times longer than $0.25$, using only delays will fail to represent the high-fidelity function due to multivaluedness (see FIG.\ref{fig:sh2}(e) in section \ref{sec:phase}); $t$ itself used as an observable will resolve this.

\subsection{Extending the formulation through delays}
\label{sec:extend}

Now we provide a mathematical formulation of the approach.
We assume that $f_h:\mathbb{R}\to\mathbb{R}$ and $f_l:\mathbb{R}\to\mathbb{R}$ are $C^K$-smooth functions with $K\geq 2$, and we want to construct an interpolant of $f_h$ with a numerical scheme (here, a Gaussian Process). We assume that
\begin{enumerate}
\item[(1)] only a small number of data points $\{x,f_h(x)\}$ as well as the function $f_l$ and its $K$ derivatives are available,
\item[(2)] $f_h$ can be written in the form
\begin{equation}\label{eq:fh}
f_h(x)=g(x,f_l(x),f_l^{(1)}(x),\dots,f_l^{(K)}(x)),
\end{equation}
where $f_l^{(i)}(x)$ denotes the $i$-th derivative, $i\in\{1,\dots,K\}$, and
\item[(3)] $g:\mathbb{R}^{K+2}\to\mathbb{R}$ is a $C^K$ function with derivatives bounded in the $L^\infty$ norm by a small constant $c_g>0$,
\begin{equation}\label{eq:bounded derivatives}
\left\|\frac{\partial}{\partial x_l}g(x_1,\dots,x_{K+1})\right\|_{L^\infty}\leq c_g\ \forall l\in\{1,\dots,K+2\}.
\end{equation}
\end{enumerate}
The Taylor series of $f_h$ reveals why assumptions (2) and (3) are required:
Assume we want to evaluate $f_h$ in a small neighborhood of a point $x_0\in\mathbb{R}$ (here, w.l.o.g. we set $x_0=0$). Then, assuming $f_h$ is smooth enough, we can write
\begin{equation}\label{eq:series}
f_h(x)=f_h(0)+\frac{\partial}{\partial x}f_h(0)x+\mathcal{O}\left(\left\|\frac{\partial^2}{\partial x^2}f_h\right\|\right).
\end{equation}
Since we assume the form (\ref{eq:fh}) for $f_h$, we can write
\begin{eqnarray}\label{eq:derivatives1}
\frac{\partial}{\partial x}f_h(x)&=&\frac{\partial}{\partial x}\left[ g(x,f_l(x),f_l^{(1)}(x),\dots,f_l^{(K)}(x))\right]\\\label{eq:derivatives2}
&=&\frac{\partial}{\partial x_1}g(x,f_l(x),f_l^{(1)}(x),\dots)+\\\label{eq:derivatives3}
&&\sum_{i=2}^{K+1}\frac{\partial}{\partial x_i}g(x,f_l(x),f_l^{(1)}(x),\dots)\cdot f_l^{(i-1)}(x).
\end{eqnarray}
From this, we can see that $\frac{\partial}{\partial x}f_h(x)$ can be large (because the derivatives $f_l^{(i)}(x)$ can be large), but if we know all $f_l^{(i)}(x)$, we only have to estimate $g$ from {\emph data} $f_h$  and $f_l^{(i)}$. Crucially, we do not approximate the function $f_h(x)$ and its derivatives. The derivatives of $g$ are bounded by $c_g$ through assumption (\ref{eq:fh}), and $g$ is a $C^K$ function, so only a few data points are necessary for a good fit.
If we have access to function values of $f_l$ over ``delays'' (at discrete shifts, say in the form of a  finite difference stencil) in space, rather than its derivatives, we can use the Newton series approximation of $f_h$ instead of equation (\ref{eq:series}) for an analogous argumentation: For functions $f_h$ that are analytic around the expansion point (here, $x_0=0$), the function $f_h$ can be evaluated at $x$ close to it by
\begin{eqnarray}
f_h(x)&=&\sum_{m=0}^\infty\left(\begin{matrix}x-m\\m\end{matrix}\right)\Delta_m f_h(0)=\\&&f_h(0)+\frac{f_h(\Delta x)-f_h(0)}{\Delta x}x  +\mathcal{O}(\|\Delta_2 f_h\|),
\end{eqnarray}
where $\Delta_m f_h$ is the $m$-th finite difference approximation of $f_h$, with a small step size $\Delta x$. By equation (\ref{eq:fh}), these differences can be expressed through $g$ and delays of $f_l$ (instead of delays of $f_h$), analogously to equations (\ref{eq:derivatives1}--\ref{eq:derivatives3}). Using delays in space compared to derivatives has numerical advantages, especially in cases where $f_h$ or $f_l$ are not differentiable (or even have discontinuities). It also enables us to estimate the derivatives of $f_l$ implicitly, in case only the function $f_l$ is available (and not its derivatives).

\subsection{Outline of the numerical approach}

In order to employ the delay coordinates of the low-fidelity function, it is required to know shifts of it. 
A necessary condition of the proposed framework is that the low-fidelity function is given explicitly or can be {\em well-learned} by given data such that low-fidelity function values can be accurately approximated (interpolated) at arbitrary points. 
If the state variable $t$ is not available, the low fidelity model should be a generic observation of $t$ to be useful in employing Takens' embedding theorem~\cite{sauer91,Takens81}. 
Under these conditions, we now present a summary of the workflow.

If the low-fidelity model is given in the form of (rich) data, we train a GP regression model for it from these data $\{\bold{t_l},\bold{y_l(t_l)}\}$ via minimizing a negative log marginal likelihood estimation. 
This data driven process can be circumvented if the low-fidelity model is explicitly given, as in the above examples.
After that, we compute predictive posterior means of the low-fidelity model at the points $t_h$ where the high-fidelity data is available.
We also compute a number of shifts of the low-fidelity function at the points $t_h-k\tau$ and at the test points $t^*$.
Next, we train another GP regression model for high-fidelity data in the higher-dimensional space, $\{\{\bold{t_h},\bold{f_l(t_h)}, \bold{f_l(t_h -\tau)}, \dots , \bold{f_l(t_h -n\tau)}\}, \bold{f_h(t_h)}\}$. 
Finally, we compute the predictive posterior mean and variance at the test points ($t^*$) in the higher-dimensional space. 
Each new delay burdens the optimization by a single additional hyperparameter.
For more details, refer to~\cite{rasmussen06, perdikaris17}. 
In this paper, all GP computations are performed by the open source python package GPy~\cite{gpy14}.

\section{Results}
\label{sec:result}
We introduce three pedagogical examples to demonstrate our approach.
First, we explore the case where $f_h$ is a phase shifted version of $f_l$ (section \ref{sec:phase}).
Then, we show that oscillations with different periods (leading to different recurrences) present a more challenging scenario, which however
can still be resolved by using shifts of $f_l$. 
The third example involves discontinuities in $f_h$ and $f_l$.
After these three examples, in section \ref{sec:HH}, we demonstrate the approach in the context of the Hodgkin-Huxley model.
We investigate the effectiveness of the proposed framework by comparing it to three established frameworks: (1) single Gaussian Process regression with high-fidelity data only (GP or Kriging), (2) Auto-regressive method with a constant scaling parameter $\rho$ (AR1 or coKriging), and (3) nonlinear auto-regressive GP (NARGP) in the same computational environment. 

\subsection{Phase Shifted Oscillations}
\label{sec:phase}
\begin{figure*}
 \centering
  \subfigure[The correlation between models]{
  \includegraphics[scale=0.33]{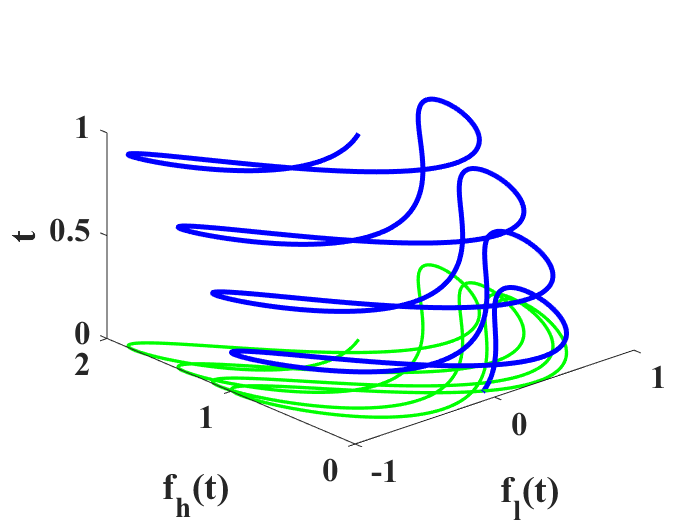}
   }
 \subfigure[GP (Kriging)]{
  \includegraphics[scale=0.4]{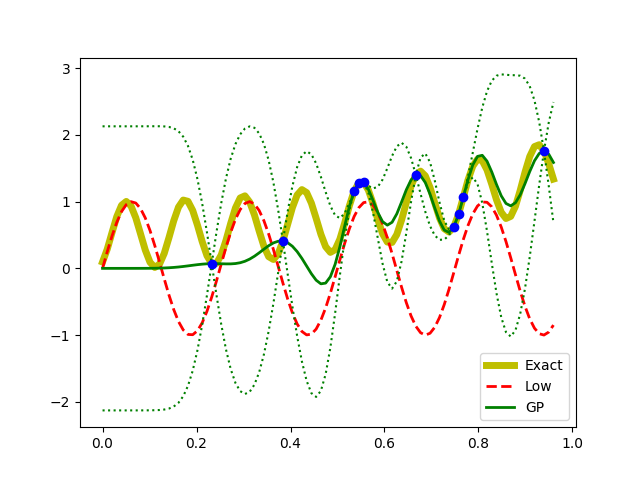}
   }
 \subfigure[AR1 (CoKriging)]{
  \includegraphics[scale=0.4]{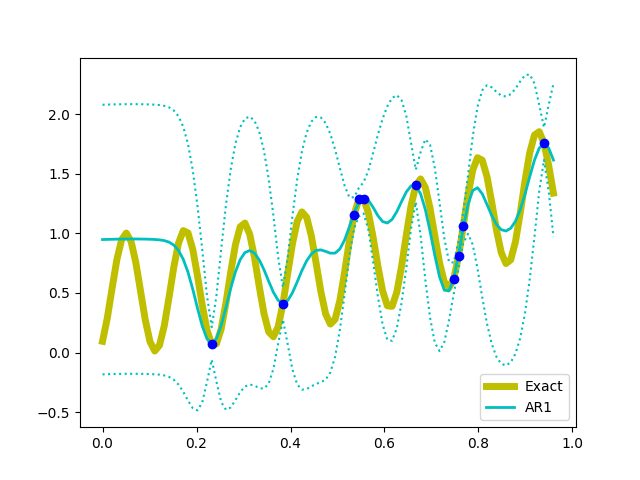}
   }
 \subfigure[NARGP]{
  \includegraphics[scale=0.4]{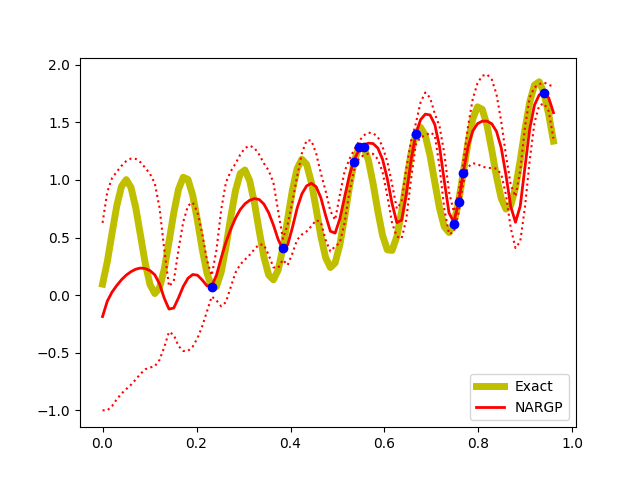}
   }
   \subfigure[GP+delays]{
  \includegraphics[scale=0.4]{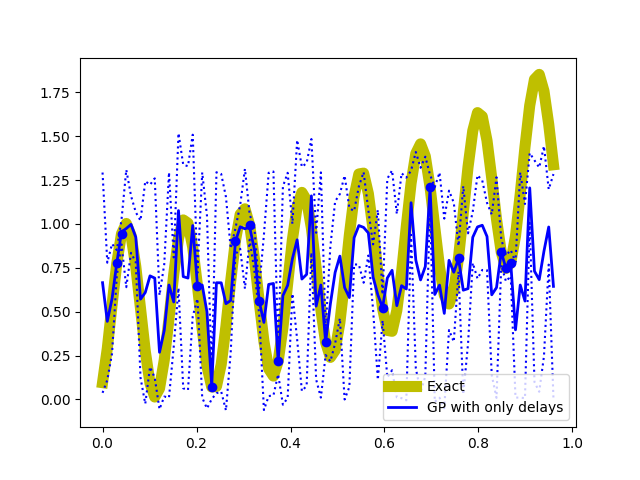}
   }
 \subfigure[GP+E]{
  \includegraphics[scale=0.4]{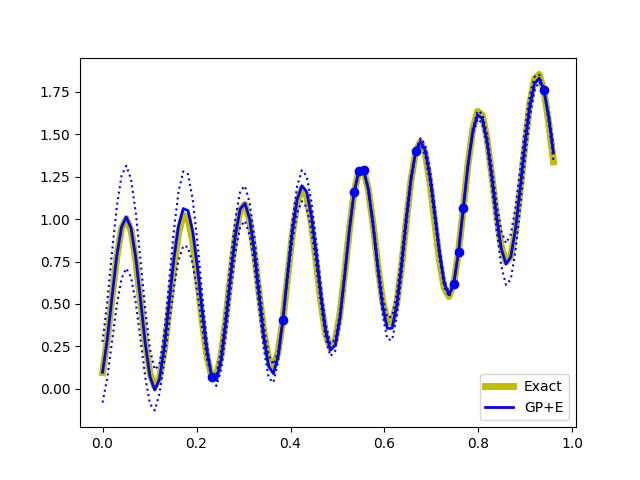}
   }

  \caption{Examples of phase shifted oscillations. (a) correlation between $t$, $f_l$ and $f_h$. (b)-(f) The high-fidelity function (the yellow curve) versus posterior means with 2 standard deviations (dashed lines) of 5 compared methods with 10 high-fidelity data points and 100 uniformly distributed low-fidelity data points. (b) GP (Kriging) and low fidelity data (the red-dashed curve). (c) Auto-regressive GP (AR1 or coKriging). (d) nonlinear auto-regressive GP (NARGP). (e) GP in the higher-dimensional space with only delays $(f_l(t),f_l(t-\tau),f_l(t-2\tau))$.
  (f) GP in the higher-dimensional space (GP+E), using $(t,f_l(t),f_l(t-\tau),f_l(t-2\tau))$.} 
\label{fig:sh2}
\end{figure*}

\begin{figure*}
 \centering
  \subfigure[Phase Shifted Oscillations]{
  \includegraphics[scale=0.3]{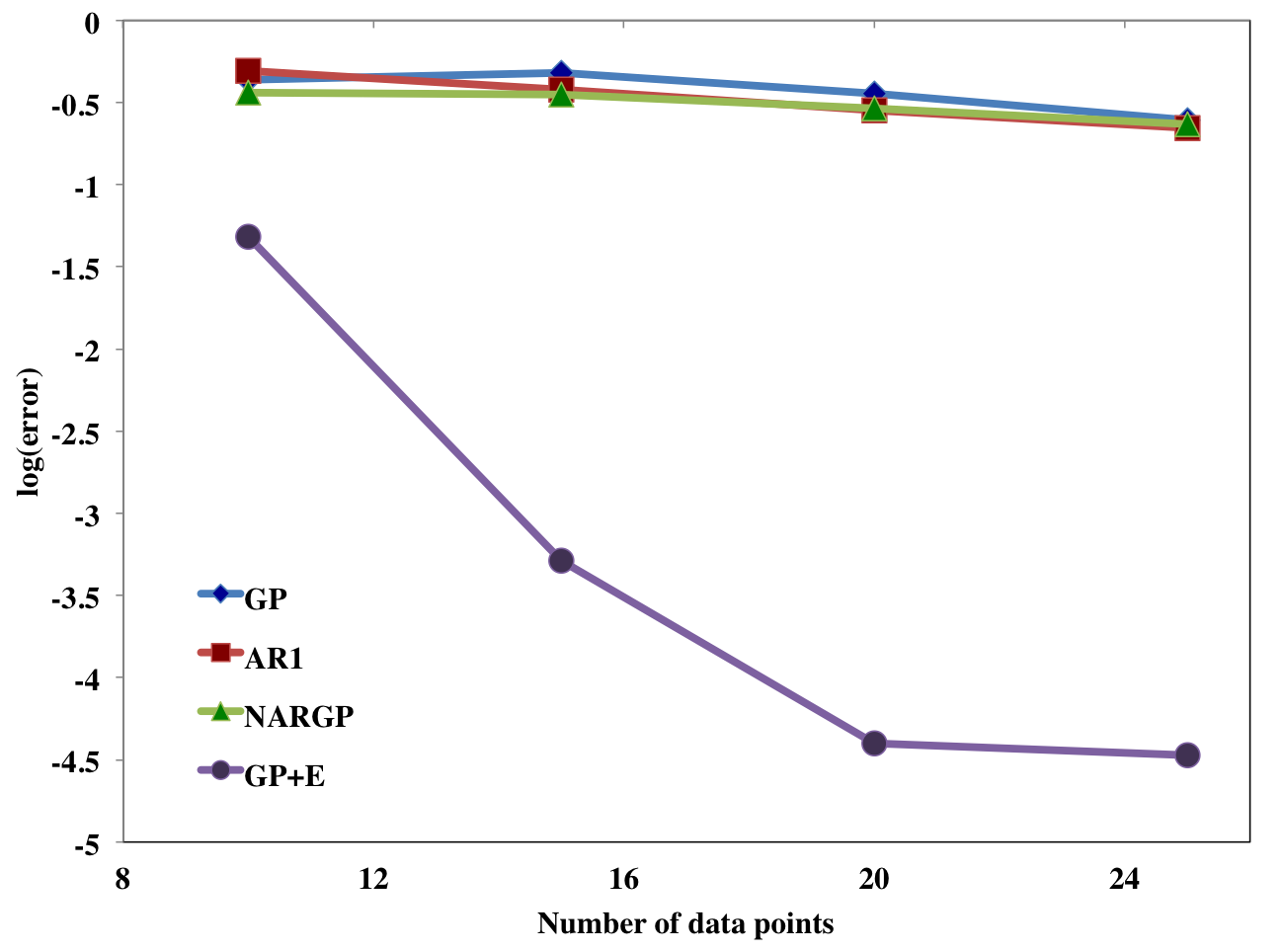}
   }
 \subfigure[Different Periodicity]{
  \includegraphics[scale=0.3]{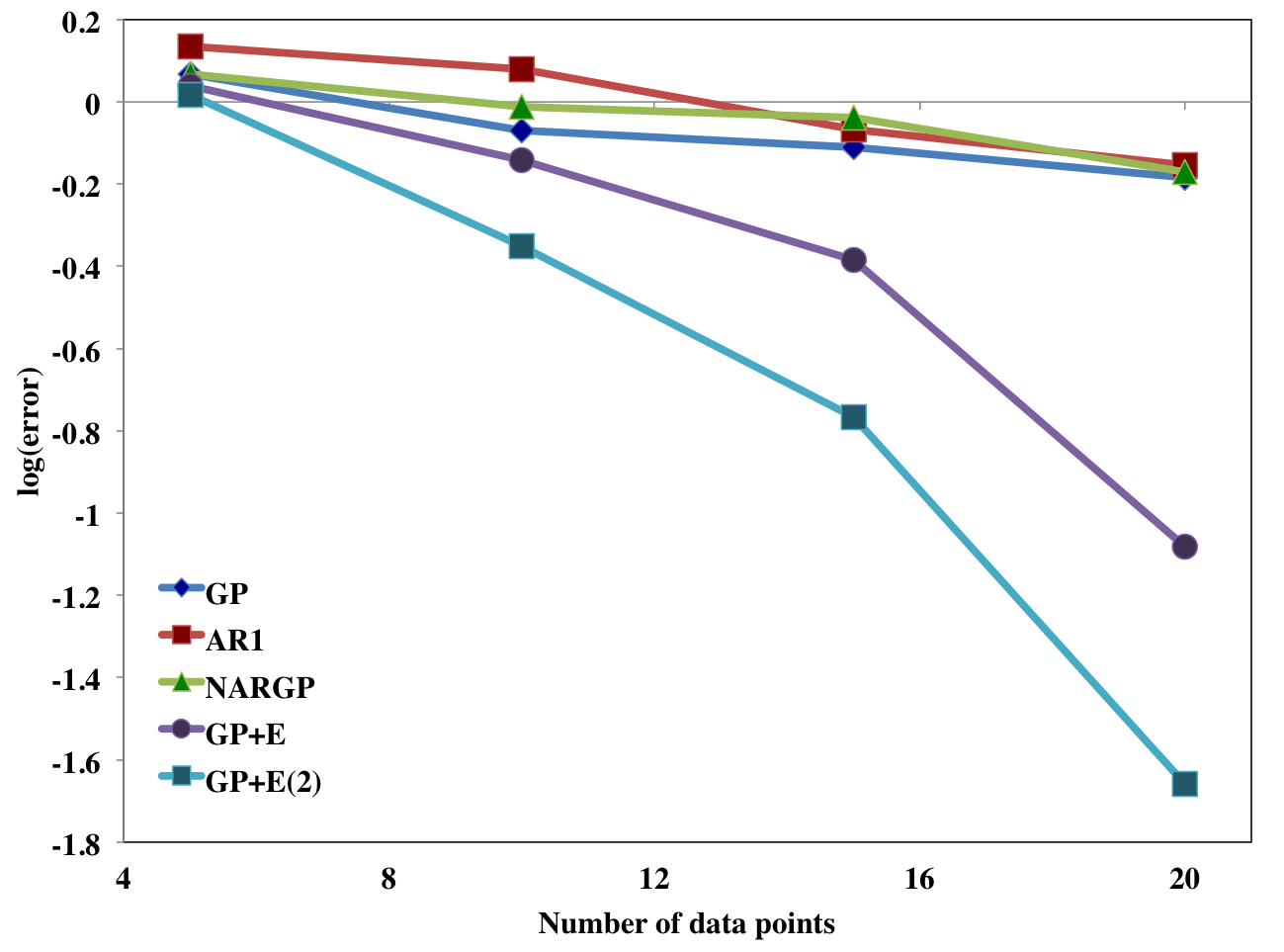}
   }
 \subfigure[Models with Discontinuities]{
  \includegraphics[scale=0.3]{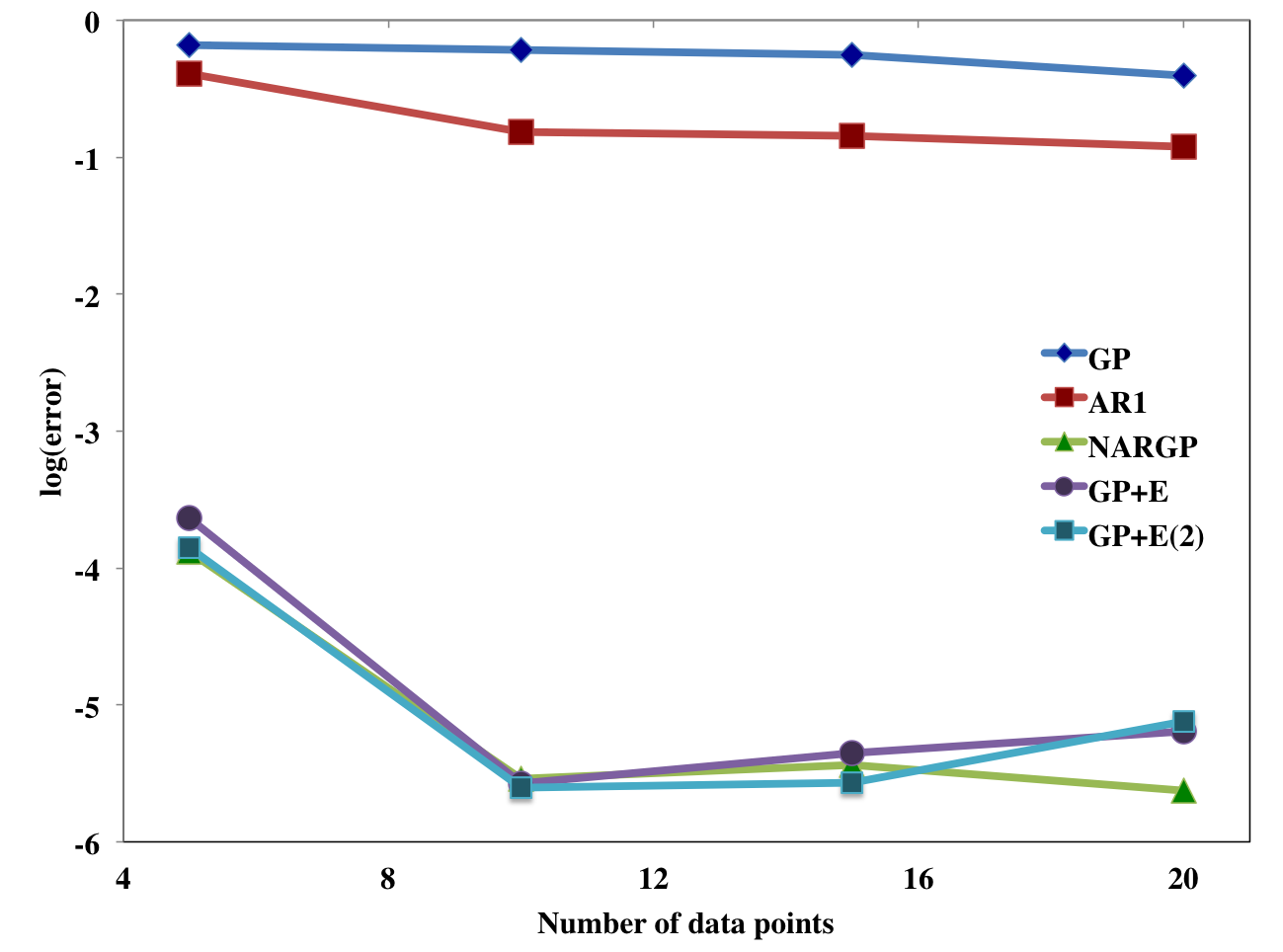}
   }
    \subfigure[The Hodgkin-Huxley models]{
  \includegraphics[scale=0.3]{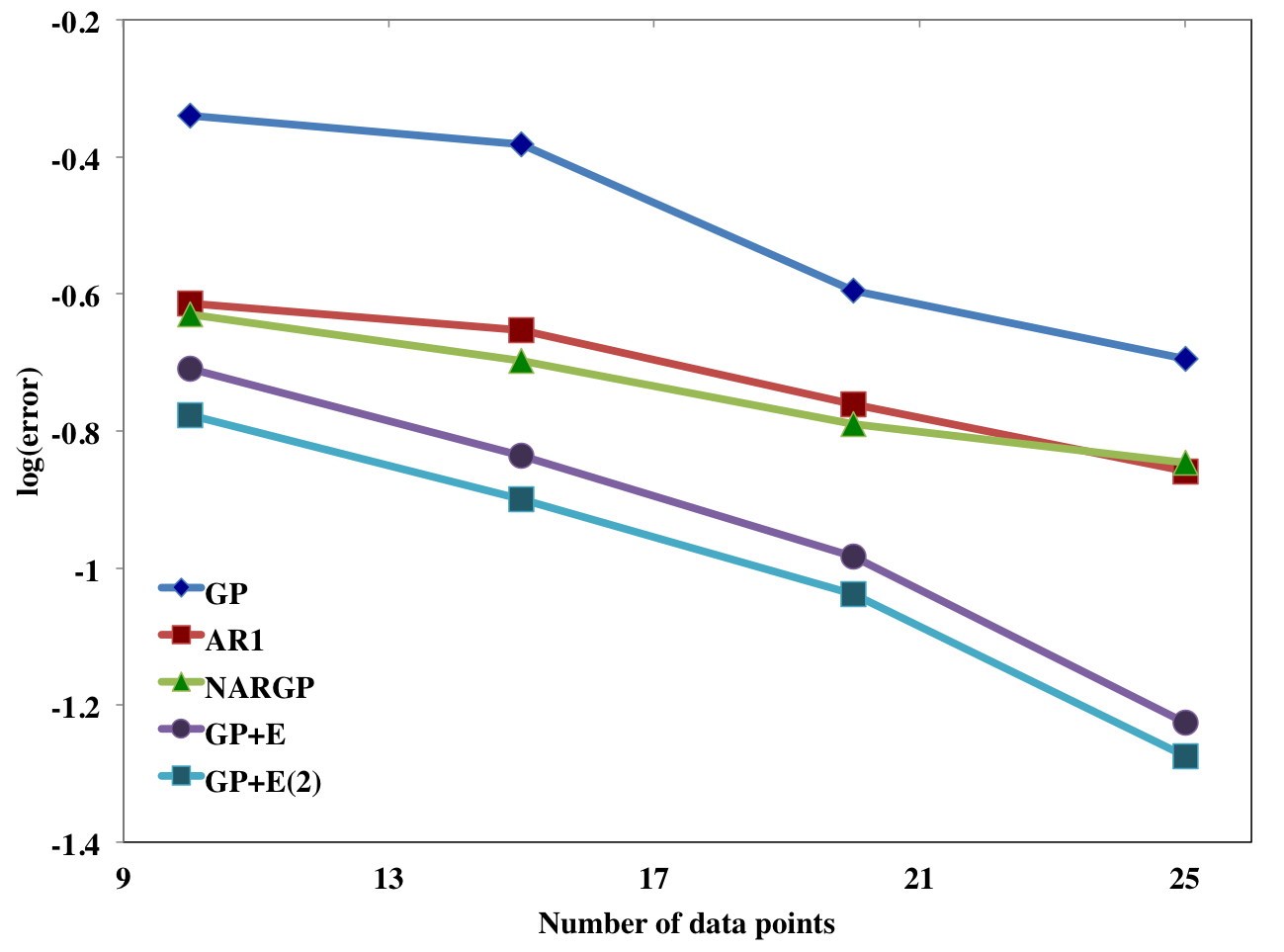}
   }
  \caption{Log $L_2$ error (y-axis) of prediction by GP (Kriging), AR1 (coKriging), NARGP, and GPs in the higher-dimensional space (GP+E and GP+E(2)) with respect to the number of high-fidelity data points (x-axis). 
The error is obtained by averaging 10 trials of random data selections.} 
\label{fig:sen}
\end{figure*}

Using models at different levels of resolution (e.g. for biological oscillators) will often give oscillations that have very comparable periods but are phase-shifted.
Let us start by considering two functions with different phases on $t\in[0,1]$,
\begin{equation}
f_h(t) = t^2 + \sin^2 (8\pi t + \pi/10), \;\;\; f_l(t) = \sin (8\pi t).
\end{equation}

The high-fidelity function can be rewritten by a trigonometric addition formula:
\begin{equation}
f_h(t) = t^2 + (\sin (8\pi t)\cos(\pi/10) + \cos (8\pi t)\sin(\pi/10))^2.
\end{equation} 
Now we can explicitly see how the high-fidelity function can be thought of as a combination of three variables: $t^2$, the low-fidelity function $f_l(t) = \sin(8\pi t)$ and its first derivative $f_l^{(1)}(t) = \cos(8\pi t)$.

Using delays of $f_l$ with a small stepsize $\tau$ contains enough information to numerically estimate its derivatives, hence we can also write $f_h$ as 
\begin{equation}
\label{eqn:g1}
f_h(t) \mapsto g(t,f_l(t),f_l(t-\tau),f_l(t-2\tau)).
\end{equation}
The GP regression model for $g$ is trained on this 4-dimensional data. %
In addition, we perform GP regression in a three dimensional space constructed from only three delays:
\begin{equation}
f_h(t) \mapsto g(f_l(t),f_l(t-\tau),f_l(t-2\tau)).
\end{equation}

As shown in FIG.\ref{fig:sh2}(b), the single GP regression model provides inaccurate predictive posterior means due to lack of high-fidelity data.
While the linear auto-regressive model (AR1) also fails to predict the high fidelity values, the NARGP (with 10 high-fidelity data points and 100 low-fidelity data points) catches the trend of the high fidelity data, yet still yields inaccurate results: NARGP is informed only by $t$ and $f_l(t)$, and not also by $f_l^{(1)}(t)$.
Similarly, the GP regression with only delays (no information about $t$) in FIG.\ref{fig:sh2}(e) fails to represent the high-fidelity function for these long observation windows.
Beyond $0.25$, $t$ cannot be recovered from the shifts of $f_l$ 
because $f_l$ is only a generic observer of $t\in[0,0.25]$.

As shown in FIG.\ref{fig:sh2}(f), the GP using $t$ and three delays of $f_l$ provides an excellent prediction with only 10 high-fidelity data points (and 100 low-fidelity data points). 
This means that, in the 4-dimensional space, $g$ (see equation \ref{eqn:g1}) has small derivatives, which then helps to employ GP regression successfully.
    
Next, we investigate the sensitivity and scalability of the proposed framework on the number of high-fidelity data points (training data points). 
We train all GP regression models with 10, 15, 20, and 25 randomly-chosen high-fidelity data points and 100 uniformly distributed low-fidelity data points. 
The error is obtained by averaging 10 trials of random data selections.
A log $L_2$ error with respect to the number of high-fidelity data points is presented in FIG.\ref{fig:sen}(a). 

Two established approaches (AR1 and NARGP) and the GP with only delays have no significant accuracy enhancements as the number of training points increases.
The reason for the consistently large errors is the lack of additional information provided by the derivatives.
The GP in the higher-dimensional space that includes $t$, on the other hand, shows a strong correlation between accuracy and the number of training points -- more high fidelity points visibly improve the approximation.

\subsection{Different Periodicity} 

In this example, the high- and the low-fidelity model oscillations are not just phase shifted, but they also are characterized by different periods.
In applications, this could arise if we tried to match observations of oscillations of {\emph the same model} at two {\emph different parameter values}. 
Different (possibly irrationally related) oscillation periods
dramatically complicate the correlation across the two data sets.

We consider two different period {\emph and} phase shifted data,
\begin{equation}
f_h(t) = \sin (8\pi t +\pi/10), \;\;\;\; f_l(t) = \sin (6\sqrt{2}\pi t).
\end{equation}

The high-fidelity function can be rewritten by a trigonometric addition formula:
\begin{equation}
f_h(t) = \sin (8\pi t)\cos(\pi/10) + \cos (8\pi t)\sin(\pi/10).
\end{equation}
In addition, the first term $\sin(8\pi t)$ can be rewritten again by a trigonometric subtraction formula:
\begin{equation}
\sin(at-bt) = \cos(bt)\sin(at) - \cos(at)\sin(bt),
\end{equation}
where $a=6\sqrt{2}\pi$ and $b=6\sqrt{2}\pi-8\pi$.
Then, 
\begin{equation}
\sin(8\pi t) = \cos(bt)f_l(t) - \sin(bt)f_l^{(1)}(t).
\end{equation}
The second term $\cos(8\pi t)$ can be rewritten in the same way.
This shows that the high-fidelity function can be written in terms of $\sin(bt)$, $\cos(bt)$, $f_l(t)$, and $f_l^{(1)}(t)$. 
Since $\sin(bt)$ and $\cos(bt)$ have lower frequency compared to the original frequency $8$, the bound $c_g$ for the derivatives of $g$ (see section \ref{sec:extend}) is smaller. 
It is then reasonable that we can approximate the high-fidelity function in the higher-dimensional space with a few training data points.

\begin{figure*}
 \centering
  \subfigure[The correlation between models]{
  \includegraphics[scale=0.3]{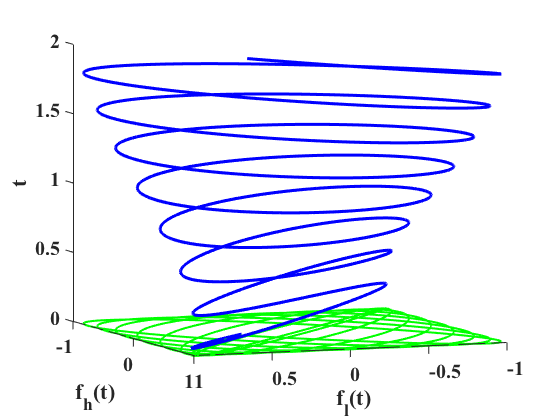}
   }
 \subfigure[GP (Kriging)]{
  \includegraphics[scale=0.3]{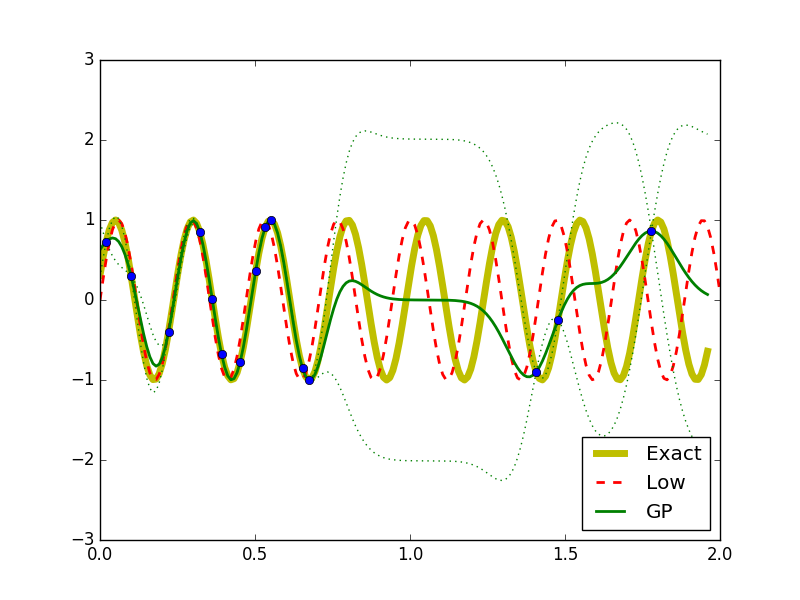}
   }
 \subfigure[AR1 (CoKriging)]{
  \includegraphics[scale=0.3]{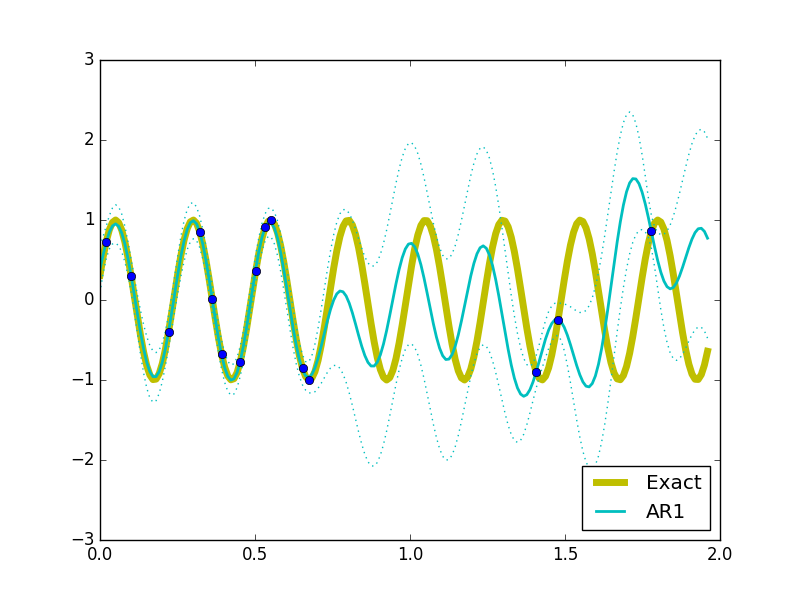}
   }
 \subfigure[NARGP]{
  \includegraphics[scale=0.3]{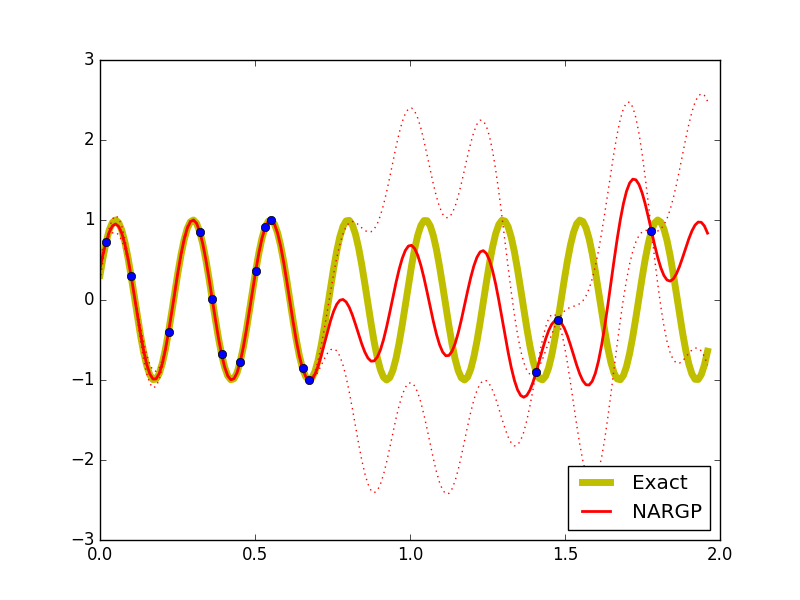}
   }
  \subfigure[GP+E]{
  \includegraphics[scale=0.3]{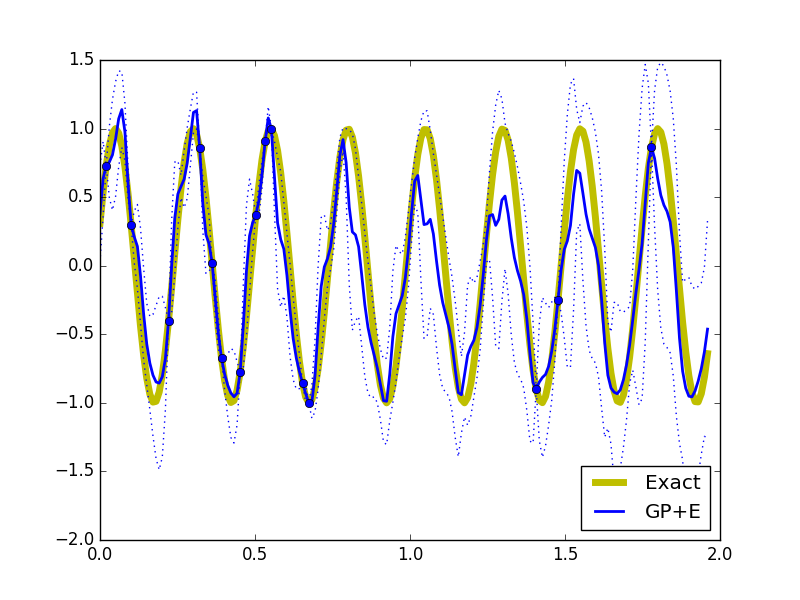}
   }
 \subfigure[GP+E(2)]{
  \includegraphics[scale=0.3]{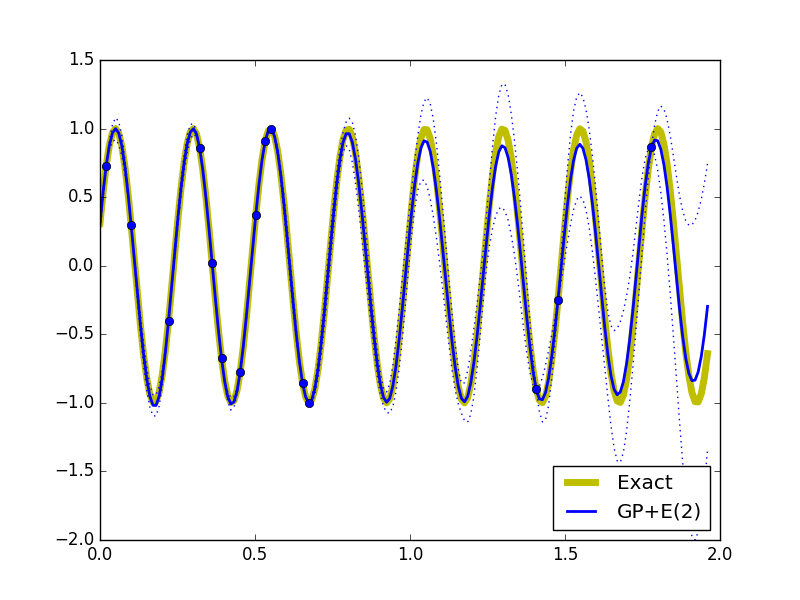}
   } 
  \caption{Examples of different periodicity. (a) correlation between $f_l$ and $f_h$. (b)-(f) The high-fidelity function (the yellow curve) versus posterior means with 2 standard deviations (dashed lines) of 5 compared methods with 15 high-fidelity data points and 200 uniformly distributed low-fidelity data points. (b) GP (Kriging) and low fidelity data (the red-dashed curve). (c) Auto-regressive GP (AR1 or coKriging). (d) nonlinear auto-regressive GP (NARGP). (e) GP in the 4-dimensional space (GP+E), using $(t, f_l(t),f_l(t-\tau),f_l(t-2\tau))$.
  (f) GP in the 6-dimensional space (GP+E(2)), using $(t,f_l(t),f_l(t-\tau),f_l(t-2\tau),f_l(t-3\tau),f_l(t-4\tau))$.} 
\label{fig:par}
\end{figure*}

We perform the GP in the two different extended spaces: (1) 3 additional delays, totaling 4-dimensional space (GP+E), and (2) 5 additional delays, totaling 6-dimensional space (GP+E(2)), and compare them to a single GP, AR1, and NARGP. 
Examples of regression results with 15 high-fidelity data points and 200 uniformly distributed low-fidelity data are shown in FIG.\ref{fig:par}. 
The GP in the 4-dimensional space provides  better regression result than other established methods, and the GP in the 6-dimensional space presents the best results.

Moreover, as shown in FIG.\ref{fig:par}(b), the phase discrepancy between the high- and low-fidelity functions increases as time increases, resulting in larger error for larger values of $t$, see FIG.\ref{fig:par}(b)--(d).
However, the GPs in the higher-dimensional spaces provide accurate prediction results over this time observation window.

The sensitivity of the number of high fidelity data is shown in FIG.\ref{fig:sen}(b). 
The GPs in the 4- and 6-dimensional space show significant computational accuracy gain compared to all other methods. 
These results demonstrate the capability of the proposed framework for period-shifted and phase-shifted data fusion.

\subsection{Models with Discontinuities}

\begin{figure*}
 \centering
  \subfigure[The correlation between models]{
  \includegraphics[scale=0.33]{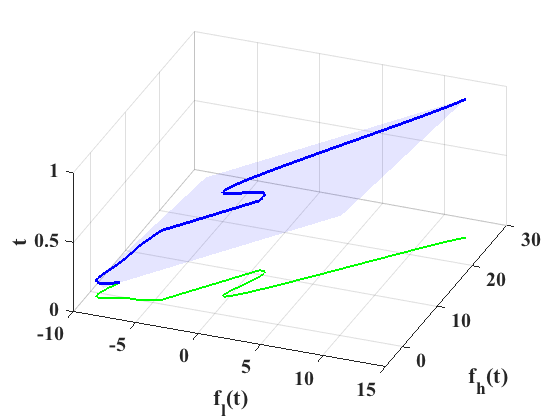}
   }
 \subfigure[GP (Kriging)]{
  \includegraphics[scale=0.4]{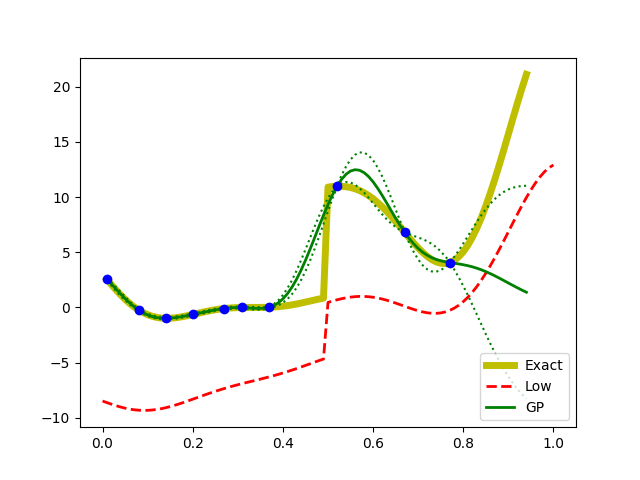}
   }
 \subfigure[AR1 (CoKriging)]{
  \includegraphics[scale=0.4]{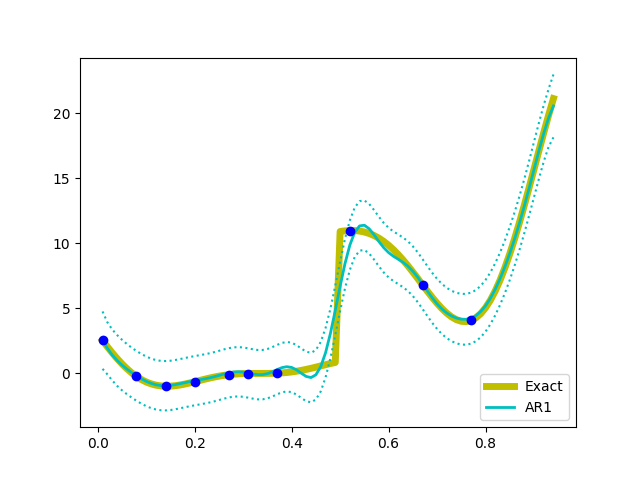}
   }
 \subfigure[NARGP]{
  \includegraphics[scale=0.4]{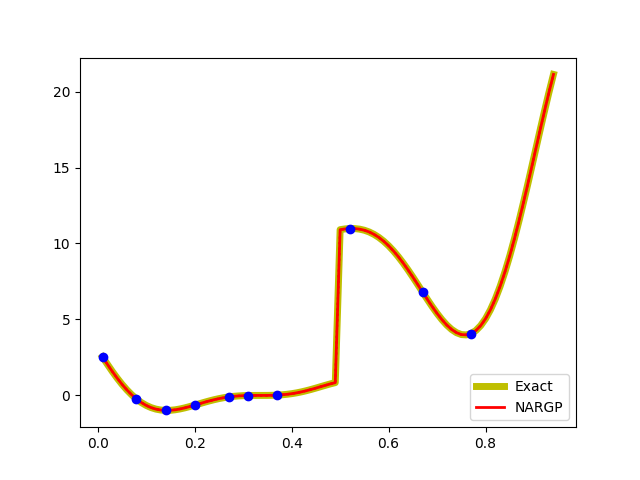}
   }
  \subfigure[GP+E]{
  \includegraphics[scale=0.4]{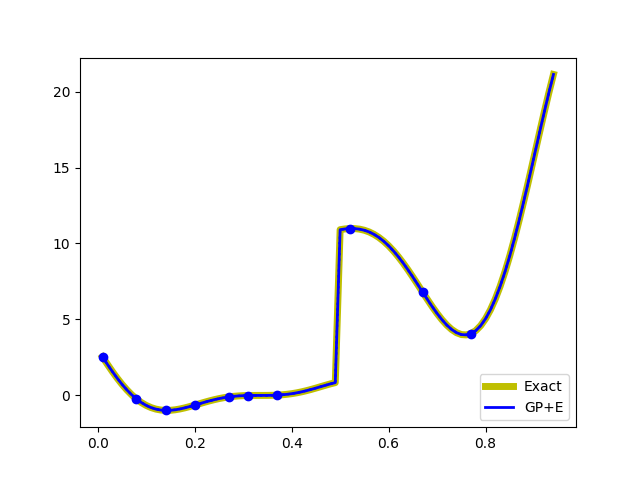}
   }
 \subfigure[GP+E(2)]{
  \includegraphics[scale=0.4]{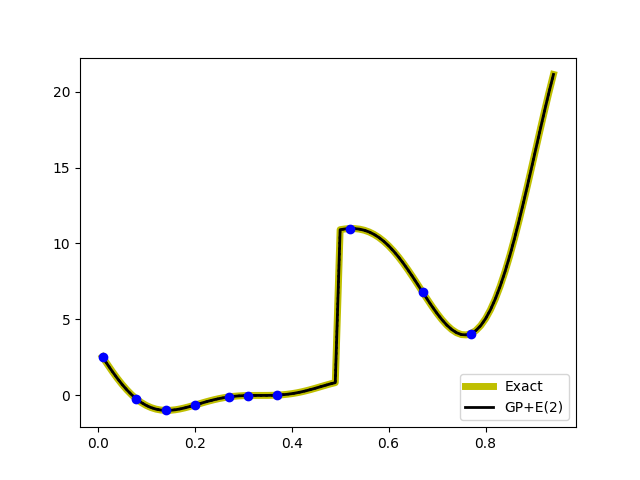}
   } 
  \caption{Examples of models with discontinuities. (a) correlation between $t$, $f_l$ and $f_h$. (b)-(f) The high-fidelity function (the yellow curve) versus posterior means with 2 standard deviations (dashed lines) of 5 compared methods with 10 high-fidelity data points and 200 uniformly distributed low-fidelity data points. (b) GP (Kriging) and low fidelity data (the red-dashed curve). (c) Auto-regressive GP (AR1 or coKriging). (d) nonlinear auto-regressive GP (NARGP). (e) GP in the 4-dimensional space (GP+E), using $(t, f_l(t),f_l(t-\tau),f_l(t-2\tau))$.
  (f) GP in the 6-dimensional space (GP+E(2)), using $(t,f_l(t),f_l(t-\tau),f_l(t-2\tau),f_l(t-3\tau),f_l(t-4\tau))$.} 
\label{fig:dis}
\end{figure*}

In general, a smooth stationary covariance kernel cannot capture discontinuous model data.
In order to resolve this problem, a non-stationary kernel has been introduced~\cite{schmidt03, paciorek04}, with space-dependent hyperparameters.
Moreover, nested GPs were also used successfully to alleviate this problem~\cite{hensman14}. 
Both approaches introduce, of course, additional hyperparameters to optimize.

In this example, we introduce a discontinuous function $f_l$ on $t \in [0, 0.5)$,
\begin{equation}
f_l(t)=0.5(6t-2)^2\sin(12t-4)+10(t-0.5)-5,
\end{equation}
and on $t \in [0.5, 1]$ as
\begin{equation}
f_l(t)=0.5(6t-2)^2\sin(12t-4)+10(t-0.5),
\end{equation}
and the high-fidelity function $f_h$
\begin{equation}
f_h(t)=2f_l(t)-20t+20=g(t, f_l(t)).
\end{equation}
In the scenario we describe here, the high-fidelity function $f_h$ is discontinuous, but can be written in terms of a linear function of $g$ in two variables.

Examples of regression results with 10 high-fidelity data points and 200 uniformly distributed low-fidelity data points are shown in FIG.\ref{fig:dis}. 
Since $g$ is a {\em linear} function of $t$ and $f_l(t)$, the NARGP, as well as our GPs in the higher-dimensional spaces, provide highly accurate prediction results with just a few high-fidelity data.

In the sensitivity analysis of the number of high fidelity data points, see FIG.\ref{fig:sen}(c), there is no significant accuracy gain after 10 high-fidelity data points.
That is because 10 training data points are enough to represent a linear function accurately.
It is worth noting that, here, the NARGP provides better prediction results with 20 training data points compared to the GPs in the higher-dimensional space, possibly due to overfitting.

\subsection{The Hodgkin-Huxley Model}
\label{sec:HH}
\begin{figure*}
 \centering
  \subfigure[The correlation between models]{
  \includegraphics[scale=0.3]{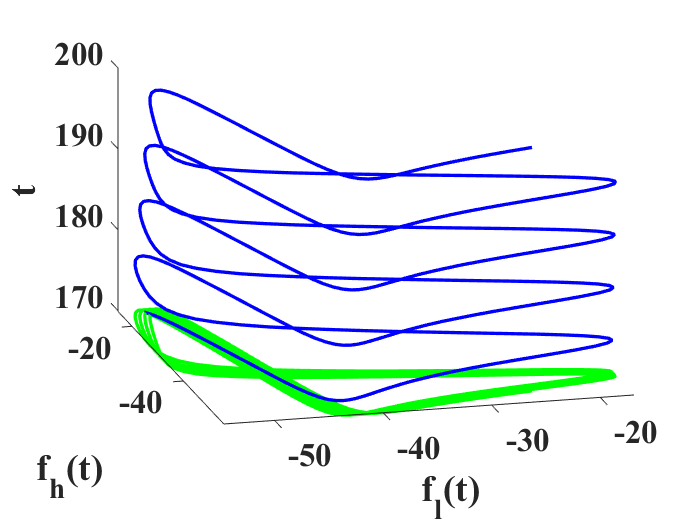}
   }
 \subfigure[GP (Kriging)]{
  \includegraphics[scale=0.4]{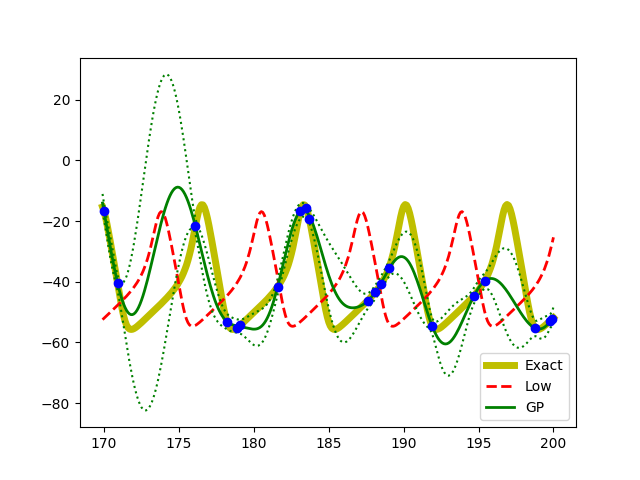}
   }
 \subfigure[AR1 (CoKriging)]{
  \includegraphics[scale=0.4]{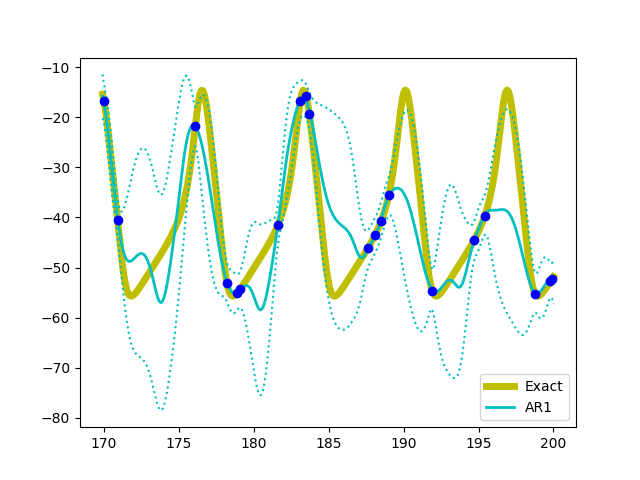}
   }
 \subfigure[NARGP]{
  \includegraphics[scale=0.4]{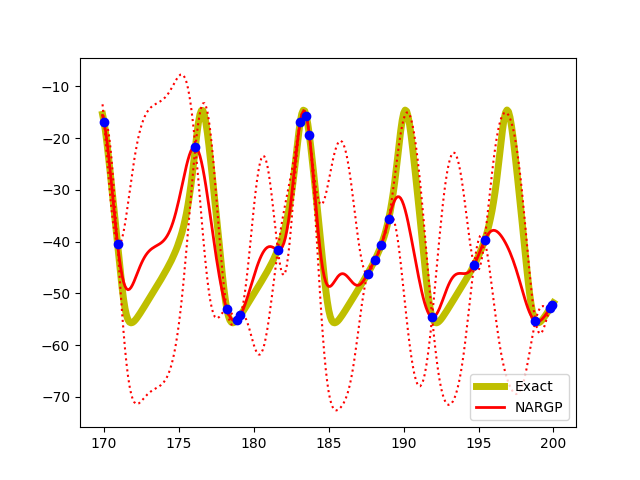}
   }
  \subfigure[GP+E]{
  \includegraphics[scale=0.4]{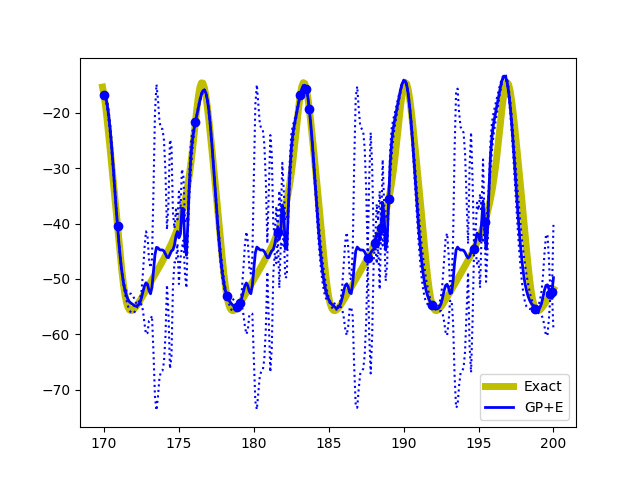}
   }
 \subfigure[GP+E(2)]{
  \includegraphics[scale=0.4]{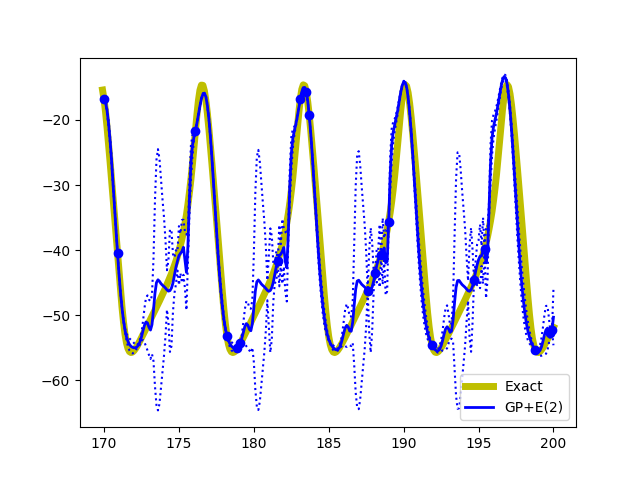}
   } 
  \caption{Examples of the two Hodgkin-Huxley model oscillations
  obtained with different external currents $I_{ext}$. (a) correlation between $t$, $f_l$ and $f_h$. (b)-(f) The high-fidelity function (the yellow curve) versus posterior means with 2 standard deviations (dashed lines) of 5 compared methods with 20 high-fidelity data points and 300 uniformly distributed low-fidelity data points. (b) GP (Kriging) and low fidelity data (the red-dashed curve). (c) Auto-regressive GP (AR1 or coKriging). (d) nonlinear auto-regressive GP (NARGP). (e) GP in the 4-dimensional space (GP+E), using $(t, f_l(t),f_l(t-\tau),f_l(t-2\tau))$.
  (f) GP in the 6-dimensional space (GP+E(2)), using $(t,f_l(t),f_l(t-\tau),f_l(t-2\tau),f_l(t-3\tau),f_l(t-4\tau))$.} 
\label{fig:HH}
\end{figure*}

Based on the results of the pedagogical examples, we apply the proposed framework to a famous model of an intracellular process, a version of the Hodgkin-Huxley equations~\cite{Hodgkin52}. 
In 1952, Hodgkin and Huxley introduced a mathematical model which can describe the initiation and propagation of action potentials in a neuron.
Specifically, they invented electrical equivalent circuits to mimic the ion channel, where ions traffic through the cell membrane.
The model for intracellular action potentials ($V_m$) can be written as a simple ODE:
\begin{equation}
C_m\frac{dV_m}{dt} + I_{ion} = I_{ext},
\label{eqn:HH}
\end{equation}
where $C_m$ is the membrane capacitance and $I_{ion}$ and $I_{ext}$ represent the total ionic current and the external current, respectively. 

The total ionic current $I_{ion} = I_{Na} + I_K + I_L$ is the sum of the three individual currents as a sodium current ($I_{Na}$), a potassium current ($I_K$), and a leakage current ($I_L$).
In order to calculate the three individual currents in time, the Hodgkin-Huxley model introduced {\em gates} which regulate the flow of ions through the channel.
Specifically, the three ionic currents are affected by the three different gates $n$, $m$, and $h$.
Based on these gates, the total ionic currents can be calculated by
\begin{equation}
I_{ion} = \bar{g}_{Na}m^3h(V_m-E_{Na})-\bar{g}_Kn^4(V_m-E_K)-\bar{g}_L(V_m-E_L),
\end{equation}
where $\bar{g}_*$ represents a normalized constant and $E_*$ represents the equilibrium potential for a sodium, $(* \equiv Na)$, a potassium, $(* \equiv K)$, and a leakage, $(* \equiv L)$, current.
The three gates $n$, $m$, and $h$ can be modeled by the following ODEs:
\begin{equation}
\frac{dn}{dt}=\alpha_n(V)(1-n)-\beta_n(V)n,
\label{eqn:n}
\end{equation}
\begin{equation}
\frac{dm}{dt}=\alpha_m(V)(1-m)-\beta_m(V)m,
\end{equation}
\begin{equation}
\frac{dh}{dt}=\alpha_h(V)(1-h)-\beta_h(V)h.
\label{eqn:h}
\end{equation}

In this paper, we set the model parameter values to $\bar{g}_{Na}=1.2$, $\bar{g}_{K}=0.36$, $\bar{g}_{L}=0.003$, $E_{Na}=55.17$, $E_{K}=-72.14$, and $E_{L}=-49.42$~\cite{Siciliano12}. 
We assume that different fidelity data come from simulations at different external currents $I_{ext}$. 
We set $I_{ext} = 1.0$ for the high fidelity model and $I_{ext} = 1.05$ for the low fidelity model, resulting in different oscillation period (and a phase shift when we start at the same initial conditions).
The action potentials $V_m$ of the two different fidelity models are shown in FIG.\ref{fig:HH}(a,b).

Examples of regression results for 5 different methods with 20 high-fidelity data points and 300 uniformly distributed low-fidelity data points are shown in FIG.\ref{fig:HH}(b--f).
Since the two data sets are phase-shifted, the single GP, AR1, and NARGP fail to represent a proper function of the high fidelity model effectively.
However, GPs in the higher-dimensional spaces provide reasonable prediction results.
The GP in the 6-dimensional space (GP+E(2)) shows significant improvement in the form of large uncertainty reduction as well as high prediction accuracy. 

The sensitivity to the number of high fidelity data is shown in FIG.\ref{fig:sen}(d). 
The proposed framework shows computational accuracy gains compared to all other methods, as well as marked improvement when new high fidelity points are added.

\section{Conclusion}
\label{sec:conclusion}   
In this paper we explored mathematical algorithms for multifidelity information fusion and its links with ``embedology", motivated by the NARGP approach.
These modifications/extensions of kriging show promise in improving the representation of data-poor ``high-fidelity" datasets exploiting data-rich ``low-fidelity" datasets. 
Given that $f_h$ may not be a simple function -and sometimes not even a function- of $f_l$, we demonstrated that using additional functions of $t$, such as derivatives or shifts of $f_l$, can drastically improve the approximation of $f_h$ through Gaussian Processes.


The limitations of the proposed framework arise in the form of the curse of dimensionality and of overfitting. 
As the number of hyperpameters in the GP framework grows in an increasingly higher dimensional input space, the optimization cost grows (and there is always the possibility of converging to local, unsatisfactory minima). Adaptively testing for the ``best" number of delays is possible, and will be
pursued in future work.  The natural option of using multiple low-fidelity models (instead of delays of just one of them) is also being explored.
Techniques that systematically find all the local hyperparameter minima (in the spirit of the reduced gradient method~\cite{quapp03}) may also be useful in this effort.
Another promising research direction involves the construction of data-informed kernels (e.g. through ``Neural-net-induced Gaussian process", NNGP~\cite{pang18}) for more realistic and unbiased predictions. Alternatively, it is interesting to consider transformations of the input space using manifold learning techniques and the so-called Mahalanobis distance~\cite{singer08,singer09}, which has been demonstrated to successfully match different (yet conjugate) models~\cite{dsilva16,kemeth18}.

What we believe is a most promising direction for the use of these techniques is the reconciliation of different granularity multiscale models - having, say, an atomistic ``high-fidelity" simulation enhanced by a continuum ``low-fidelity" approximate closure.
Thus, ``heterogeneous data" fusion becomes a version of multifidelity data fusion~\cite{lee17D}.  
In this paper the fusion tools simply ``filled in the gaps" in a single manifestation of the high fidelity data. In a time-dependent context, ``full space, full time" low fidelity simulations can help complete and thus accelerate ``small space, small time" high fidelity simulations - in a form reminiscent of the patch-dynamics approach in equation-free computation~\cite{kevrekidis03}, see also~\cite{lee17D,lee17R}. Using a qualitatively correct (even though quantitatively inaccurate) low-fidelity model --as opposed to just the local Taylor series that play the role of low-fidelity modeling in patch dynamics-- may very much improve the computational savings of such multiscale computation schemes.

\section{Acknowledgments}
    S. Lee, F. Dietrich and I. G. Kevrekidis gratefully acknowledge the partially support of NSF, NIH, and DARPA. G. E. Karniadakis gratefully acknowledge the financial support of DARPA (grant no. N66001-534 15-2-4055) also the MIT ESRDC grant.

\nocite{*}
\bibliographystyle{vancouver}
\bibliography{embedding}

\end{document}